\documentclass{acm_proc_article-sp}

\begin{document}

\title{Long-short Term Motion Feature for Action Classification and Retrieval}

\numberofauthors{1} 
\author{
\alignauthor
Zhenzhong Lan, Xuanchong Li, Ming Lin, Alexander G. Hauptmann\\
School of Computer Science, Carnegie Mellon University\\
{\tt\small lanzhzh,  xcli, minglin,alex@cs.cmu.edu}
}

\date{20 April 2013}

\maketitle
\begin{abstract}
We propose a method for representing motion information for video classification and retrieval. We improve upon local descriptor based methods that have been among the most popular and successful models for representing videos. The desired local descriptors need to satisfy two requirements: 1) to be representative, 2) to be discriminative. Therefore, they need to occur frequently enough in the videos and to be be able to tell the difference among different types of motions. To generate such local descriptors, the video blocks they are based on must contain just the right amount of motion information. However, current state-of-the-art local descriptor methods use video blocks with a single fixed size, which is insufficient for covering actions with varying speeds. In this paper, we introduce a long-short term motion feature that generates descriptors from video blocks with multiple lengths, thus covering motions with large speed variance. Experimental results show that, albeit simple, our model achieves state-of-the-arts results on several benchmark datasets. 

\end{abstract}




\section{Introduction}

\begin{figure}
\centering
\includegraphics[height = 2cm,width=8.3cm]{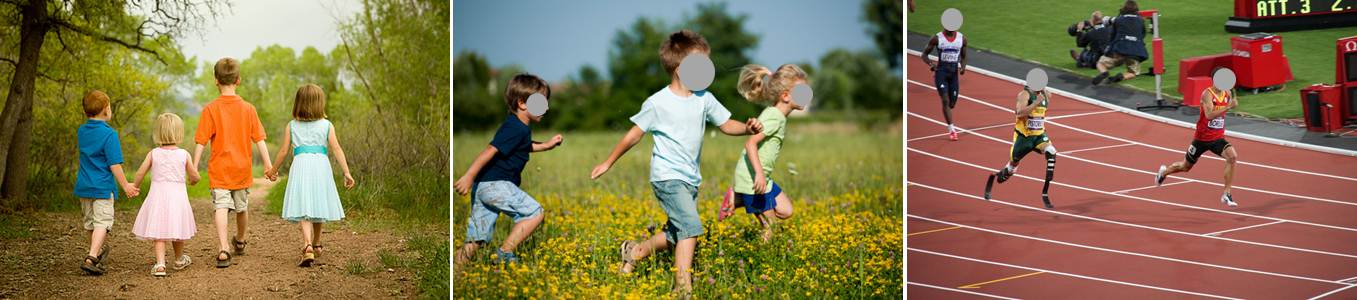}
\caption{Between action classes (``walking'' and ``running''), different actions can have different speed. But even within each class (``running''), the speed of different action can be dramatically different due to different performers and performing time. }
\label{fig:illustration_action}
\end{figure}

With the explosive growth of the user generated on-line videos and the prevailing on-line video sharing communities, content-based video retrieval \cite{jiang2007towards, amir2003ibm, snoek2008concept, lan2013cmu} has become an important problem in multimedia retrieval. Because of the large visual diversity of on-line videos, robust video representations become the key component for solving this problem. Among them, local spatio-temporal features have been the most popular and successful methods for representing videos. A local spatio-temporal feature is computed in 3 steps: (1) extracting fixed sized local video blocks, i.e., cubiod or trajectory; (2) describing local video blocks, i.e., using Histogram of Optical Flow (HOF) and/or Motion Boundary Histogram (MBH); (3) encoding and pooling local video descriptors, i.e. using Bag of Features (BoF) or Fisher Vector. This paper focuses on improving the first step which aims to find the right video primitives for representing motion information. What are the right primitives for representing motion information? This is a fundamental question that has been asked for several decades. At one extreme, it can be a pixel but there is not enough information for a pixel to make a discriminative descriptor. At the other extreme, it can be a whole video but that is too specific to be generalizable. As a consequence, state-of-art methods use video blocks with one fixed size. For example, \cite{wang2013action} uses trajectories across 15 frames while \cite{simonyan2014two} generates features from a sequence of 10 frames. However, unconstrained on-line videos often contain actions that have large speed differences, as illustrated in Figure \ref{fig:illustration_action}. Video blocks with single size would have difficulty in covering actions with large speed differences. Slower motions require longer time to finish hence longer video blocks to generate discriminative descriptors while faster motions need short video blocks to be represented. Although long-term video block is generally more discriminative \cite{simonyan2014two}, it is also less representative and harder to get due to the difficulty of tracking \cite{wang2013action}.

To handle this difficulty, we propose a long-short term motion feature (LSTMF) that pools features from multiple video blocks with different lengths. LSTMF relies on the idea that with multiple block sizes, actions have a higher chance to find the right block size hence right description for them. Although, it is quite a simple idea, it can be used as a powerful, indiscriminately applicable tool that can be adopted by any video description methods. Our experimental results on several benchmark datasets also show that state-of-the-arts performance can be achieved if we combine LSTMF with Improved Dense Trajectory (IDT) \cite{wang2013action}.

In the remainder of this paper, we provide more background information about video retrieval and motion features. We then describe LSTMF and its application on IDT in detail. After that, an evaluation of our method is performed and a comparison of the results with other state-of-art methods is given. We conclude with a discussion of our method.

\section{Related Work}

Video retrieval research has been largely driven by the advances of video representation methods \cite{jiang2007towards, amir2003ibm, snoek2008concept, lan2013cmu}. There is an extensive body of literature about video representations; here we just mention a few relevant ones involved with state-of-the-art feature extractors and feature encoding methods.  See \cite{aggarwal2011human} for an in-depth survey. 

Most traditional video representation methods are based on high-dimensional encodings of local spatio-temporal features. For instance, Space-time Interest Points (STIP) \cite{laptev2005space} consists of detecting video cuboids, which are then described using histogram of gradient (HOG)  and histogram of optical flow (HOF). The features are then encoded in a BoF manner, which aggregates features over several spatio-temporal grids. More recently, the Dense Trajectory method proposed by Wang et al. \cite{wang2011action,wang2013action}, together with the Fisher Vector encoding  \cite{perronnin2010improving}  yields the current state-of-the-art performances on several benchmark action recognition datasets. . Peng et al. \cite{peng2014bag} further improved the performance of Dense Trajectory by increasing the codebook sizes and fusing multiple coding methods. Some success has been reported recently using deep convolutional neural networks for action recognition in videos. Karpathy et al. \cite{karpathy2014large} trained a deep convolutional neural network using 1 million weakly labeled YouTube videos and reported a moderate success using it as a feature extractor. Simonyan $\&$ Zisserman \cite{simonyan2014two} reported a result that is competitive to IDT \cite{wang2013action} by training deep convolutional neural networks using both sampled frames and optical flows.

\section{ Long-short Term Motion Feature (LSTMF)}

We now formalize our model. Given a video $\mathcal{V}$, we first do video block extraction: \\$\Phi(\mathcal{V}): \mathcal{V} \rightarrow \{b_1(\phi_1,w,h,l), b_2(\phi_2,w,h,l),..., b_n(\phi_n,w,h,l)\}$. $\phi_i$ are a $3 \times l$ matrices, in which each column is a 3-tuple indicating the space-time location of the video block. $(w, h, l)$ are the width, height and length of the video block, respectively.  Since we only focus on the length of the video block, we omit $(w,h)$ in further discussion and denote a video block as $b_i(\phi_i,l)$. Traditionally, all the $b_i$ share the same fixed $l$. For example, for STIP $l=2$, for dense trajectory $l=15$, for two-stream Convolutional Networks $l=10$.  In LSTMF, we allow each $b_i$ have different $l$. That is, $b_i=b_i(\phi_i,l_i)$. We denote function $g: b_i \rightarrow \mathbb{R}^D$ as a local descriptor generator such as SIFT and $f: g(b_i) \rightarrow \mathbb{R}^K$ as the encoding and pooling function. Based on those definitions, we express the long-short term feature of video $\mathcal{V}$ as 
\begin{align}
X(\mathcal{V}) = f(g(b_1(\phi_1,l_1)),g(b_2(\phi_2,l_2)),...,g(b_n(\phi_n,l_M)))
\label{f1}
\end{align}

\section{Experiments}
We examine our proposed LSTMF representation on several video retrieval tasks, predominately involving actions. The experimental results show that LSTMF representations outperform conventional representations with single-size video blocks on these difficult real-world datasets. 

\subsection{Experimental Setting}
We use IDT with Fisher Vector encoding \cite{wang2013action} to evaluate our method because it represents the current state-of-the-arts for most real-world action recognition datasets. 

We use the same settings as in \cite{wang2013action} for our baseline. These settings include the IDT feature extraction, Fisher vector representation and a linear SVM classifier.

IDT features are extracted using 15 frame tracking, camera motion stabilization with human masking and RootSIFT \cite{arandjelovic2012three} normalization and described by Trajectory, HOG, HOF and MBH descriptors. PCA is used to reduce the dimensionality of these descriptors by a factor of two. 

For Fisher vector representation,  we map the raw feature descriptors into a  Gaussian Mixture Model with 256 Gaussians trained from a set of randomly sampled 256000 data points. Power and L2 normalization are also used before concatenating different types of descriptors into a video based representation. 

For classification, we use a linear SVM classifier with a fixed C=100 as recommended by \cite{wang2013action} and the one-versus-all approach is used for multi-class classification scenario. 

For LSTMF, besides $l = 15$, we also add $l=30, 45, 60, 75$ and $90$. Since the Trajectory descriptor size is based on the video block length, we subsample the Trajectory descriptor to match the $l=15$ Trajectory descriptor length. Because descriptors for longer video black can be constructed from the descriptors of shorter video block, calculating descriptors for longer videos incurs almost no additional computational cost.

\subsection{Datasets}

\begin{figure*}
\centering
\begin{tabular}{cc}
\fbox{\includegraphics[height = 2cm, width=6.6cm]{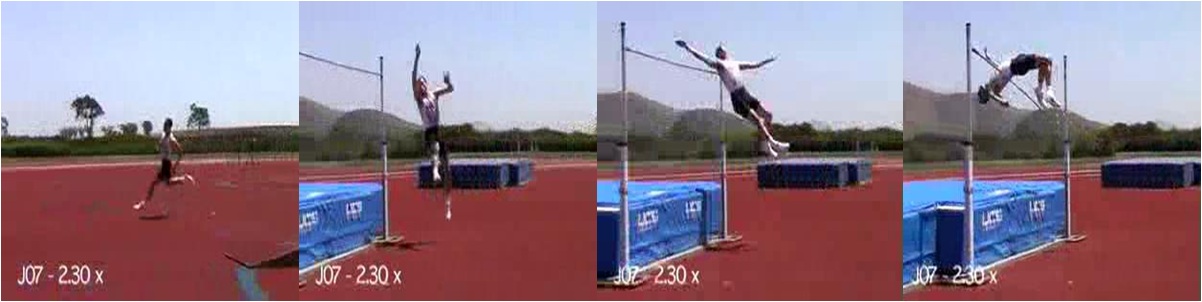}}&
\fbox{\includegraphics[height = 2cm,width=6.6cm]{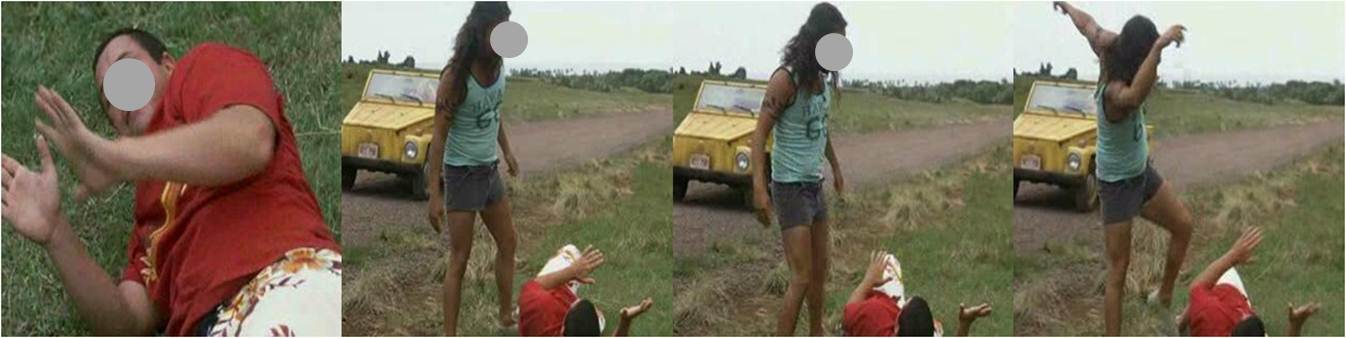}}\\
(a) HighJump &(b) Kick\\
\fbox{\includegraphics[height = 2cm, width=6.6cm]{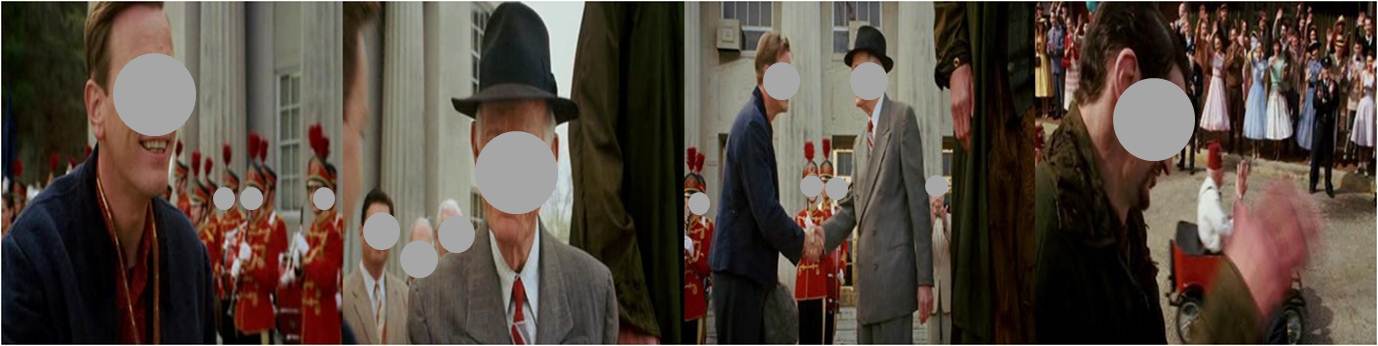}}&
\fbox{\includegraphics[height = 2cm,width=6.6cm]{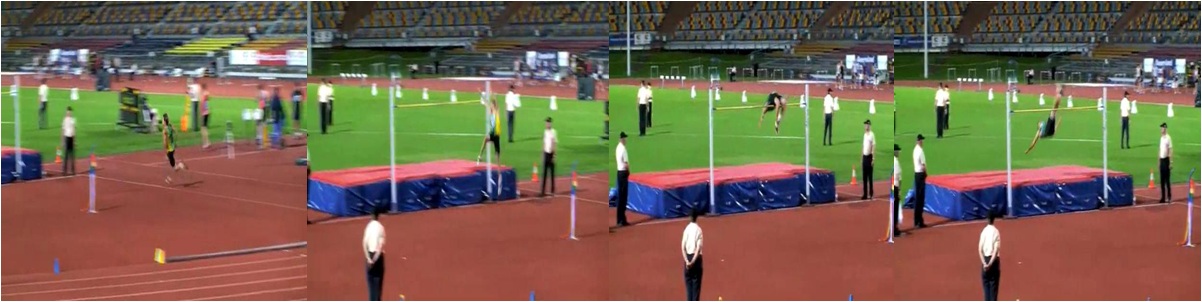}}\\
(c) HandShake&(d)HighJump
\end{tabular}
\caption{Examples frames from (a) UCF50, (b) HMDB51, (c) Hollywood2, (d) Olympic Sports.}
\label{fig:examples}
\end{figure*}

\begin{table}
\centering
\begin{tabular}{|c |c |c | }
\hline
 Datasets  & Avg. \# Samples& Avg. Durations (s) \\
   &   (Train/Test)&  \\\hline
UCF50 &128/5 & 7.44  \\
HMDB51  &70/30 & 3.14 \\
Hollywood2  & 67/74 & 11.55  \\
Olympic  & 41/8 & 7.74  \\
\hline
\end{tabular}
\caption{\label{tab:dataset}Meta data of the experimental datasets.}
\end{table}

We use four video retrieval or classification datasets, UCF50, HMDB51, Hollywood2 and Olympic Sports, for evaluation. Example frames are shown in Fig. \ref{fig:examples}. 
These datasets, which mainly involve actions, are selected because they are the real-world action datasets that have received the bulk of experimental attention.

The UCF50 dataset \cite{reddy2013recognizing} has 50 action classes spanning over 6618 YouTube videos clips that can be split into 25 groups. The video clips in the same group are generally very similar in background. Leave-one-group-out cross-validation as recommended by \cite{reddy2013recognizing} is used and mean accuracy (mAcc) over all classes and all groups is reported. 

The HMDB51 dataset \cite{kuehne2011hmdb} has 51 action classes and 6766 video clips extracted from digitized movies and YouTube. \cite{kuehne2011hmdb} provides both original videos and stabilized ones. We only use original videos in this paper and standard splits with mAcc are used to evaluate the performance. 

The Hollywood2 dataset \cite{marszalek2009actions} contains 12 action classes and 1707 video clips that are collected from 69 different Hollywood movies. We use the standard splits with training and test videos provided by \cite{marszalek2009actions}.   Mean average precision (mAP) is used to evaluate this dataset because multiple labels can be assigned to one video clip. 

The Olympic Sports dataset \cite{niebles2010modeling} consists of 16 athletes
practicing sports, represented by a total of 783 video clips. We use standard splits with 649 training clips and 134 test clips and report mAP as in \cite{niebles2010modeling} for comparison purposes.  Note that as shown in Table \ref{tab:dataset}, in this standard split, each class only has about 8 testing samples, so the results of this dataset may not be able to reliably evaluate the quality of the model.

\subsection{Experimental Results}

\subsubsection{Results of LSTMF}
In Table \ref{tab:scalePyramid}, we list both single-length and LSTMF performance.  We first examine how performance changes with respect to $l$ (single-length). We can see that as we increase $l$, in all of the datasets except Olympic sports, the performance decreases. This result is consistent with Wang \& Schimld \cite{wang2013action} and mostly because they have already picked the optimal trajectory length for these datasets. It also demonstrates that the performance of a feature greatly relies on the choice of the block size hence the importance of finding the right video block size for a local descriptor based model. We also observe that the performance of HMDB51 decreases dramatically. This decrease is because the average duration of HMDB51 videos are significantly shorter than videos in other datasets, as shown in Table \ref{tab:dataset}. As we increase $l$, there is a large portion of videos that generate no features.  

Next, let us check the behavior of LSTMF. We evaluate LSTMF in a pyramid manner. That is, we combine the features from all previous sets. Therefore, the results of, for example, $l=45$, is based on using the features from $l=15, l=30$ and $l=45$. We observe that for LSTMF representations, although there is small fluctuation, the performance generally increases as the value of $l$ increases. The exception is Olympics sports dataset, in which the trend is quite irregular. Again, it is worth mentioning that the improvement of Olympics sports dataset is very unreliable due to its small number of testing samples. We conjecture that the fluctuation is because longer trajectories have a higher chance to drift from the initial position \cite{wang2011action}. Overall, LSTMF IDT performs better than single-length IDT.

\begin{table*}
\centering
\begin{tabular}{|c | c |c |c | c| c|c |c |c | }
\hline
  & \multicolumn{2}{|c|}{HMDB51  } & \multicolumn{2}{|c|}{ Hollywood2 }   & \multicolumn{2}{|c|}{UCF50  } & \multicolumn{2}{|c|}{ Olympics Sports } \\
    & \multicolumn{2}{|c|}{ (MAcc$\%$) } & \multicolumn{2}{|c|}{(MAP$\%$)}  & \multicolumn{2}{|c|}{(MAcc$\%$) } & \multicolumn{2}{|c|}{  (MAP$\%$)} \\    \hline
  $l$ & single-length & LSTMF & single-length & LSTMF & single-length & LSTMF & single-length & LSTMF\\\hline 

 15 &   62.1 &                   & 67.0 &                & 93.0 &                   & 89.8 & \\
 30 &   61.7 & 62.5              & 66.8 & 67.7           & 92.9 & 93.5              & 90.0 & 91.2 \\
 45 &   54.0 & 63.2              & 66.0 & 67.5           & 92.1 & 93.4              & 89.2 & 89.8 \\
 60 &   44.5 & 63.6              & 63.9 & 68.0           & 91.0 & 93.6              & 88.0 & 91.0\\
 75 &   39.9 & \textbf{63.7}     & 61.6 & 67.8           & 87.0 & \textbf{93.8}     & 84.9 & 90.3\\
 90 &   18.0 & \textbf{63.7}     & 60.5 & \textbf{68.2}  & 81.3 & 93.7              & 61.1 & \textbf{91.4}  \\
\hline
\end{tabular}
\caption{\label{tab:scalePyramid}Comparison of LSTMFs with different $l$.}
\end{table*}

\begin{table*}
\centering
\footnotesize
\begin{tabular}{|l c |l c |l c |l c|}
\hline
    \multicolumn{2}{|c|}{HMDB51 (MAcc. $\%$)} & \multicolumn{2}{|c|}{Hollywood2 (MAP $\%$)} & \multicolumn{2}{|c|}{UCF50 (MAcc. $\%$)} & \multicolumn{2}{|c|}{Olympics Sports (MAP $\%$)}\\ \hline
    
Oneata \textit{et al.} \cite{oneata2013action}  &54.8  & Sapienz \textit{et al.} \cite{sapienza2014feature}  & 59.6    & Sanath \textit{et al.} \cite{narayan2014cause}  & 89.4  & Jain \textit{et al.} \cite{jain2013better}  & 83.2\\
Wang  \& Schmid \cite{wang2013action}   &57.2  &Jain \textit{et al.} \cite{jain2013better}    & 62.5   & Arridhana \textit{et al.} \cite{ciptadi2014movement}  & 90.0 &   Adrien \textit{et al.} \cite{gaidon2014activity}  & 85.5 \\
Simonyan \textit{et al.} \cite{simonyan2014two}  & 57.9 &  Oneata \textit{et al.} \cite{oneata2013action} &  63.3    & Oneata \textit{et al.} \cite{oneata2013action} &90.0 & Oneata \textit{et al.} \cite{oneata2013action}  & 89.0\\
Peng \textit{et al.} \cite{peng2014bag}  & 61.1 & Wang \& Schmid \cite{wang2013action}  &64.3  & Wang \& Schmid \cite{wang2013action}  & 91.2 &  Wang \& Schmid \cite{wang2013action}  & 91.1\\
\hline 
LSTMF ($l=90$)  & \textbf{63.7} & LSTMF ($l = 90$) &  \textbf{68.2} & LSTMF ($l = 90$) & \textbf{93.7}  & LSTMF ($l = 90$) &  \textbf{91.4}\\ 
\hline
\end{tabular}
\caption{\label{tab:state-of-art}Comparison of our results to the state-of-the-arts.}
\end{table*}

\subsubsection{Comparing with the State-of-the-Arts}
In Table \ref{tab:state-of-art}, we  compare LSTMF at $l=90$, with the state-of-the-art approaches. From Table \ref{tab:state-of-art}, in most of the datasets, we observe a substantial improvement over the state-of-the-arts except Olympics Sports, on which our $l=90$ LSTMF gives  marginal improvement. Note that although we list several of the most recent approaches here for comparison purposes, \textit{most of them are not directly comparable to our results due to the use of different features and representations}. The most comparable one is Wang \& Schmid \cite{wang2013action}, from which we build on our approach. Sapienz \textit{et al.} \cite{sapienza2014feature} explored ways to sub-sample and generate vocabularies for Dense Trajectory features. 
Jain et al. \cite{jain2013better}'s approach incorporated  a new motion descriptor. Oneata et al.  \cite{oneata2013action} focused on testing Spatial Fisher Vector for multiple action and event tasks. Peng et al. \cite{peng2014bag}  improved the performance of IDT by increasing the codebook size and fusing multiple coding methods. Karpathy et al. \cite{karpathy2014large} trained a deep convolutional neural network using 1 million weakly labeled YouTube videos and reported a 65.4\% mean accuracy on UCF101 datasets. Simonyan \& Zisserman \cite{simonyan2014two} reported results that are competitive to the IDT method by training deep convolutional neural networks using both sampled frames and optical flows and get a $57.9\%$ MAcc in HMDB51 and an $87.6\%$ MAcc in UCF101, which are comparable to the results of \cite{wang2013action}.

\section{Conclusion}

We propose a long-short term motion feature (LSTMF), which pools descriptors from video blocks that have different lengths. LSTMF is designed for capturing both long-term and short-term motion hence generates discriminative and representative local descriptors for unconstrained videos that have large motion variety. Experimental results on several benchmark datasets show that, although the idea is quite simple, LSTMF outperforms traditional local descriptor based methods with single-sized video blocks. In the future, we will explore having varying length video blocks for local descriptor based methods.

\bibliographystyle{abbrv}
\bibliography{sample}


\end{document}